\documentclass[%
 reprint,
 amsmath,amssymb,
 aps,
longbibliography,
nofootinbib,
]{revtex4-1}

\usepackage{graphicx}
\usepackage{dcolumn}
\usepackage{bm}
\usepackage{amsmath}
\usepackage{url}
\usepackage{hyperref}
\renewcommand{\vec}[1]{\mathbf{#1}}

\begin{document}

\title{Entity Embeddings of Categorical Variables }
\author{Cheng Guo}
 \email{cheng.guo.work@gmail.com}
\author{Felix Berkhahn}%
\email{felix.berkhahn@gmail.com}
\affiliation{%
 Neokami Inc.
}%

\date{\today}

\begin{abstract}

We map categorical variables in a function approximation problem into Euclidean spaces, which are the entity embeddings of the categorical variables. 
The mapping is learned by a neural network during the standard supervised training process.
Entity embedding not only reduces memory usage and speeds up neural networks compared with one-hot encoding,
but more importantly by mapping similar values close to each other in the embedding space it reveals the intrinsic properties of the categorical variables.
We applied it successfully in a recent Kaggle competition\footnote{\url{https://www.kaggle.com/c/rossmann-store-sales}} and were able to reach the third position with relative simple features. 
We further demonstrate in this paper that entity embedding helps the neural network to generalize better when the data is sparse and statistics is unknown. 
Thus it is especially useful for datasets with lots of high cardinality features, where other methods tend to overfit. 
We also demonstrate that the embeddings obtained from the trained neural network boost the performance of all tested machine learning methods considerably when used as the input features instead. As entity embedding defines a distance measure for categorical variables it can be used for visualizing categorical data and for data clustering.
\end{abstract}

\maketitle

\section{Introduction}
Many advances have been achieved in the past 15 years in the field of neural networks due to a combination of faster computers, more data and better methods  \cite{lecun_deep_2015}. 
Neural networks revolutionized computer vision\cite{krizhevsky_imagenet_2012, zeiler_visualizing_2014, simonyan2014very, sermanet2013overfeat, szegedy2015going}, speech recognition\cite{hinton2012deep, sainath2013deep} and natural language processing\cite{bengio_neural_2003, mikolov2011empirical, mikolov_efficient_2013, kim_convolutional_2014} and have replaced or are replacing the long dominating methods in each field. 

Unlike in the above fields where data is unstructured, neural networks are not as prominent when dealing with machine learning problems with structured data. 
This can be easily seen by the fact that the top teams in many online machine learning competitions like those hosted on Kaggle use tree based methods more often than neural networks\cite{xgboost}. 

To understand this, we compared neural network and decision tree's approach to the general machine learning problem, which is to approximate the function 
\begin{equation} 
y = f(x_1, x_2, ..., x_n) \label{fun_approx}.
\end{equation}
Given a set of input values $(x_1, x_2, ..., x_n)$ it generates the target output value $y$.

In principle a neural network can approximate any continuous function\cite{cybenko_approximation_1989, nielsen_book} and piece wise continuous function \cite{llanas2008constructive}. 
However, it is not suitable to approximate arbitrary non-continuous functions as it assumes certain level of continuity in its general form. During the training phase the continuity of the data guarantees the convergence of the optimization, and during the prediction phase it ensures that slightly changing the values of the input keeps the output stable. 
On the other hand decision trees do not assume any continuity of the feature variables and can divide the states of a variable as fine as necessary.

Interestingly the problems we usually face in nature are often continuous if we use the right representation of data. 
Whenever we find a better way to reveal the continuity of the data we increase the power of neural networks to learn the data.
For example, convolutional neural networks \cite{lecun1998gradient} group pixels in the same neighborhood together. This increases the continuity of the data
compared to simply representing the image as a flattened vector of all the pixel values of the images. 
The rise of neural networks in natural language processing is based on the word embedding \cite{bengio_neural_2003, mikolov_efficient_2013, pennington2014glove} which puts words with similar meaning closer to each other in a word space thus increasing the continuity of the words compared to using one-hot encoding of words.

Unlike unstructured data found in nature, structured data with categorical features may not have continuity at all and even if it has it may not  be so obvious.  The continuous nature of neural networks limits their applicability to categorical variables.
Therefore, naively applying neural networks on structured data with integer representation for category variables does not work well. 
A common way to circumvent this problem is to use one-hot encoding, but it has two shortcomings: First when we have many high cardinality features one-hot encoding often results in an unrealistic amount of computational resource requirement. 
Second, it treats different values of categorical variables completely independent of each other and often ignores the informative relations between them.

In this paper we show how to use the entity embedding method to automatically learn the representation of categorical features in multi-dimensional spaces which puts values with similar effect in the function approximation problem Eq.~\eqref{fun_approx} close to each other, and thereby reveals the intrinsic continuity of the data and helps neural networks as well as other common machine learning algorithms to solve the problem. 

Distributed representation of entities has been used in many contexts before\cite{hinton_learning_1986, bengio1999modeling, hinton2002learning}. 
Our main contributions are: First we explored this idea in the general function approximation problem and demonstrated its power in a large machine learning competition.
Second we studied the properties of the learned embeddings and showed how the embeddings can be used to understand and visualize categorical data.

\section{Related Work}
As far as we know the first domain where the entity embedding method in the context of neural networks has been explored is the representation of relational data\cite{hinton_learning_1986}. More recently, knowledge base which is a large collection of complex relational data is seeing lots of works using entity embedding\cite{jenatton2012latent, yang2014embedding, AAAI159342}. 
The basic data structure of relational data is triplets $(h, r, t)$, where $h$ and $t$ are two entities and $r$ is the relation.
The entities are mapped to vectors and relations are sometimes mapped to 
a matrix(e.g.\ Linear Relation Embedding \cite{paccanaro_extracting_2000}) or two matrices(e.g.\ Structured Embeddings\cite{bordes2011learning})
or a vector in the same embedding space as the entities\cite{bordes2014semantic} etc. 
Various kind of score function can be defined (see Table.~1 of \cite{he2015learning}) to measure the likelihood of such a triplet, and the score function is used as the objective function for learning the embeddings.

In natural language processing, Word embeddings have been used to map words and phrases \cite{bengio_neural_2003} into a continuous distributed vector in a semantic space.
In this space similar words are closer. 
What is even more interesting is that not only the distance between words are meaningful but also the direction of the difference vectors. For example, it has been observed \cite{mikolov_efficient_2013} that the learned word vectors have relations such as:
\begin{align}
\vec{King} - \vec{Man} \approx  \vec{Queen} - \vec{Woman}  \label{eq:word_embedding_examples1} \\
\vec{Paris} - \vec{France}  \approx \vec{Rome} - \vec{Italy}  
\label{eq:word_embedding_examples2}
\end{align}
There are many ways \cite{bengio_neural_2003, mikolov_efficient_2013, pennington2014glove, levy2014neural, kim2014convolutional} to learn word embeddings. A very fast way \cite{mikolov_distributed_2013} is to use the word context with the aim to maximize 
\begin{equation}
p(w_c|w) =\frac{\exp(\vec{w} \cdot \vec{w_c})}{\sum_i \exp(\vec{w} \cdot \vec{w_i})}, 
\end{equation}
where $\vec{w}$ and $\vec{w_c}$ are the vector representation of a word $w$ and its neighbor word $w_c$ inside the context window while $p(w_c|w)$ is the probability to have $w_c$ in the context of $w$. 
The sum is over the whole vocabulary. Word embeddings can also be learned with supervised methods. 
For example in Ref.~\cite{kim2014convolutional} the embeddings can be learned using text with labeled sentiment.
This approach is very close to the approach we use in this paper but in a different context.

\section{Tree based methods}
As tree based methods are the most widely used method for structured data and they are the main methods we are comparing to, we will briefly review them here.
Random Forests and in particular Gradient Boosted Trees have proven their capabilities in numerous recent Kaggle competitions \cite{xgboost}.
In the following, we will briefly describe the process of growing a single decision tree used for regression, as well as two popular tree ensemble methods: random forests and gradient tree boosting. 
\subsection{Single decision tree}
\label{sec:decision_tree}
Decision trees partition the feature space $X$ into $M$ different sub-spaces $R_1, R_2, \dots R_M$. The function $f$ in equation (\ref{fun_approx}) is thus modeled as
\begin{equation}
f(x) = \sum_{m = 1}^M c_m I(x \in R_m)
\end{equation}
with $I$ being the indicator function \newline
$I(x \in R_m)=\begin{cases}1&\text{if } x \in R_m \\0& \text{else} \end{cases}$.
Using the common sum of squares
\begin{equation} \label{eq:sos}
L = \sum_i \left( y_i - f(x_i) \right)^2
\end{equation}
as loss function, it follows from standard linear regression theory that, for given $R_m$, the optimal choices for the parameters $c_m$ are just the averages
\begin{equation} \label{eq:optimal_c}
\hat{c}_m = \frac{1}{|R_m|} \sum_{x_i \in R_m} y_i
\end{equation}
with $|R_m|$ the number of elements in the set $R_m$. Ideally, we would try to find the optimal partition $\{R_m\}$ such as to minimize the loss function (\ref{eq:sos}). However, this is not computationally feasible, as the number of possible partitions grows exponentially with the size of the feature space $X$. Instead, a greedy algorithm is applied, that tries to find subsequent splits of $X$ that try to minimize (\ref{eq:sos}) locally at each split. To start with, given a splitting variable $j$ and a split point $s$, we define the pair of half-planes
\begin{eqnarray}
R_1(j, s) &=& \{X | X_j \leq s \} \\
R_2(j, s) &=& \{X | X_j > s \} 
\end{eqnarray}
and optimize (\ref{eq:sos}) for $j$ and $s$:
\begin{equation} \label{eq:greedy_optimization}
\min_{j, s} \left[ \sum_{x_i \in R_1(j, s)} (y_i - \hat{c}_1)^2 + \sum_{x_i \in R_2(j, s)} (y_i - \hat{c}_2)^2  \right]
\end{equation}
The optimal choices for the parameters $\hat{c}_1$ and $\hat{c}_2$ follow directly from (\ref{eq:optimal_c}).

After (\ref{eq:greedy_optimization}) is solved for $j$ and $s$, the same algorithm is applied recursively on the two half-planes $R_1$ and $R_2$ until the tree is fully grown.

The size up to which the tree is grown governs the complexity of the model and thus implies a bias-variance tradeoff: A very large tree likely overfits the training data, while a very small tree likely is not complex enough to capture the important dependencies in the data. There are several strategies and measures available to control the tree size. A very popular strategy is \textit{pruning}, where first large trees are grown until they reach a minimal tree size (like minimum number of nodes or minimal height), and then internal nodes are collapsed (i.e. pruned) to minimize a cost-complexity measure $C_\alpha$ such as 
\begin{equation}
C_\alpha = \sum_i (y_i - f(x_i))^2 + \alpha |T|
\end{equation}
where $|T|$ is the number of terminal nodes in the tree $T$ and $\alpha$ is a free parameter to control the complexity of the model.

\subsection{Random forests}
A single decision tree is a highly non-linear classifier with typically low bias but high variance. Random forests address the problem of high variance by establishing a committee (i.e. average) of identically distributed single decision trees. 

To be precise, random forests contain $N$ single decision trees grown by the following algorithm:
\begin{enumerate}
\item Draw a bootstrap sample from the training data, that is, select $n$ random records from the training data.
\item Grow a single decision tree $T_i$ as described in section \ref{sec:decision_tree}, with the only difference that at each split-node $m$ features are randomly picked that are considered for the best split at the split-node. 
\item Output the ensemble of all decision trees $\{T_i\}_{i = 1 \dots N}$.
\end{enumerate}
For regression, an unseen sample is then predicted as:
\begin{equation} \label{eq:rf_prediction}
f(x) = \frac{1}{N} \sum_{i=1}^{N}  T_i(x)
\end{equation}
As all $T_i$ are identically distributed, the linear average of (\ref{eq:rf_prediction}) preserves the presumably low bias of a single decision tree. However, averaging will reduce the variance of the single decision trees.

\subsection{Gradient boosted trees}
Gradient tree boosting is another ensemble tree based method, that is we try to approximate $f(x)$ by a sum of trees $T_i$:
\begin{equation} \label{eq:sum_trees}
f(x) = \sum_{k=1}^N T_k(x)
\end{equation}
For a generic loss function $L$ (not necessarily quadratic), the $n$-th tree is grown on the quantity $r_{in}$ 
\begin{equation} \label{eq:loss_gradients}
r_{in} = - \frac{\partial L(y_i, f(x_i))}{\partial f(x_i)} \Bigr|_{f = f_{n-1}}
\end{equation}
computed using its $n-1$ predecessor trees. Here, the $y_i$ are the target labels, $x_i$ are the sample features and $f_{n-1}$ is the sum of the first $n-1$ trees
\begin{equation}
f_{n-1}(x) = \sum_{k=1}^{n-1} T_k(x)
\end{equation}
In case of a squared error loss $L = \sum_i \left( y_i - f(x_i) \right)^2$ this amounts to fitting the $n$-th tree on the residuals $y_i - f_{n-1}(x_i)$ of its $n-1$ predecessor trees. Hence, equation (\ref{eq:loss_gradients}) generalizes to a generic loss function by minimizing the loss function $L$ iteratively at each step along the gradient descent direction in the space spanned by all possible trees $T_n$. This is where the name \textit{gradient} boosted trees comes from.

As for every boosting algorithm, the next iterative classifier $T_n$ tries to correct its $T_{n-1}$ predecessors. Hence, in contrast to random forests, gradient tree boosting also aims to minimize the bias of the ensemble and not only the variance.

\section{Structured data}

By structured data we mean data collected and organized in a table format with columns representing different features (variables) or target values and rows representing different samples. We focus on this type of data in this paper.

The most common variable types in structured data are continuous variables and discrete variables. 
Continuous variables such as temperature, price, weight can be represented by real numbers. 
Discrete variables such as age, color, bus line number can be represented by integers. 
Often the integers are just used for convenience to label the different states and have no information in themselves. 
For example if we use 1, 2, 3 to represent red, blue and yellow, one can not assume that "blue is bigger than red" or "the average of red and yellow are blue"  or anything that introduces additional information based on the properties of integers. 
These integers are called nominal numbers.
Other times there is an intrinsic ordering in the integer index such as age or month of the year. 
These integers are called cardinal number or ordinal numbers. Note that the meaning or order may not be more useful for the problem than only considering the integer as nominal numbers.
For example the month ordering has nothing to do with number of days in a month (January is closer to Jun than February regarding number of days it has). Therefore we will treat both types of  discrete variables in the same way. The task of entity embedding is to map discrete values to a multi-dimensional space where values with similar function output are close to each other.

\section{Entity embedding}
To learn the approximation of the function Eq.~\eqref{fun_approx}
we map each state of a discrete variable to a vector as 
\begin{equation}
e_i: x_i \mapsto  \mathbf{x}_i
\end{equation}

\begin{figure}
	\centering
	\includegraphics[width=0.5\textwidth]{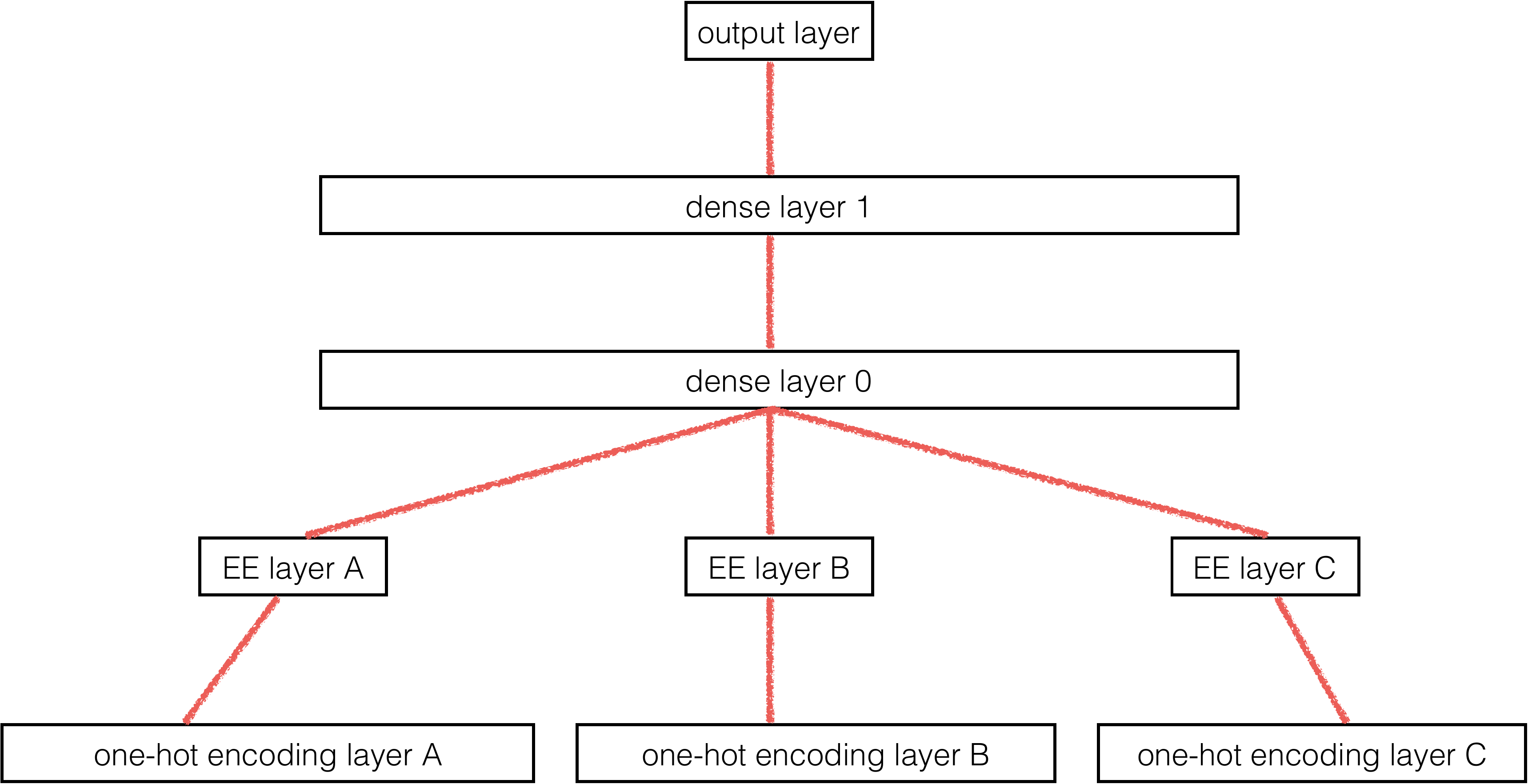}
	\caption{Illustration that entity embedding layers are equivalent to extra layers on top of each one-hot encoded input.}
	\label{fig:ee_and_onehot}
\end{figure}

This mapping is equivalent to an extra layer of linear neurons on top of the one-hot encoded input as shown in Fig.~\ref{fig:ee_and_onehot}.
To show this we represent one-hot encoding of $x_i$ as
\begin{equation}
u_i: x_i \mapsto  \delta_{x_i\alpha},
\end{equation}
where $\delta_{x_i\alpha}$ is Kronecker delta and the possible values for $\alpha$ are the same as $x_i$.
If $m_i$ is the number of values for the categorical variable $x_i$, then $\delta_{x_i\alpha}$ is a vector of length $m_i$, where the element is only non-zero when $\alpha = x_i$.

The output of the extra layer of linear neurons given the input $x_i$ is defined as
\begin{equation}
\mathbf{x}_i \equiv \sum_\alpha w_{\alpha\beta}   \delta_{x_i\alpha} = w_{x_i\beta} \label{eq:ee_onehot}
\end{equation}
where $ w_{\alpha\beta}$ is the weight connecting the one-hot encoding layer to the embedding layer and
$\beta$ is the index of the embedding layer. 
Now we can see that the mapped embeddings are just the weights of this layer and can be learned in the same way as the parameters of other neural network layers.

After we use entity embeddings to represent all categorical variables, all embedding layers and the input of all continuous variables (if any) are concatenated. The merged layer is treated like a normal input layer in neural networks and other layers can be build on top of it. The whole network can be trained with the standard back-propagation method. In this way, the entity embedding layer learns about the intrinsic properties of each category, while the deeper layers form complex combinations of them.

The dimensions of the embedding layers $D_i$ are hyper-parameters that need to be pre-defined.
The bound of the dimensions of entity embeddings are between 1 and $m_i - 1$  where $m_i$ is the number of values for the categorical variable $x_i$.  In practice we chose the dimensions based on experiments. The following empirical guidelines are used during this process: First, the more complex the more dimensions. We roughly estimated how many features/aspects one might need to describe the entities and used that as the dimension to start with. Second, if we had no clue about the first guideline, then we started with $m_i - 1$. 

It would be good to have more theoretical guidelines on how to choose $D_i$. We think this probably relates to the problem of embedding of finite metric space, and that is what we want to explore next.

\subsection{Relation with embedding of finite metric space}
\label{sec:finite_metric_space}
With entity embedding we want to put similar values of a categorical variable closer to each
other in the embedding space. 
If we use a real number to define similarity of the values then entity embedding is closely related to the 
embedding of finite metric space problem in topology.

We define a finite metric space $(\mathit{M_i}, d_i)$ associated with each categorical variable $x_i$ in the function approximation problem Eq.~\eqref{fun_approx},
where $\mathit{M_i}$ is the set of all possible values of $x_i$. $d_i$ is the metric on $\mathit{M_i}$, which is the distance function between any two pairs of values $(x_i^p, x_i^q)$ of $x_i$. 
We want $d_i$ to represent the similarity of $(x_i^p, x_i^q)$. There are many ways to define it, one simple and natural way is
\begin{equation}
	d_i(x_i^p, x_i^q) = \langle |f(x_i^p, \mathbf{\bar{x}_i}) - f(x_i^q, \mathbf{\bar{x}_i}) | \rangle_\mathbf{\bar{x}_i} \label{eq:metric}
\end{equation}
where $\langle \dots \rangle_\mathbf{\bar{x}_i}$ is the average over all values of the parameters of $f$ other than $x_i$. $\mathbf{\bar{x}_i}$ is shorter notation for $(x_1, x_2, \dots, x_{i-1}, x_{i+1}, \dots)$.
It can be verified that the following conditions hold for the metric Eq.~\eqref{eq:metric}:
\begin{align}
	&d_i(x_i^p, x_i^q) = 0 \Leftrightarrow x_i^p = x_i^q \label{eq:indentity} \\ 
	&d_i(x_i^p, x_i^q) = d_i(x_i^q, x_i^p) \\
	&d_i(x_i^p, x_i^r) \le d_i(x_i^p, x_i^q) + d_i(x_i^q, x_i^r)
\end{align}
Eq.~\eqref{eq:indentity} may not automatically hold in a real problem when two different values always generate the same output. However, this also means one value is redundant, and it is easy to simply merge these two values into one by redefining the categorical variable to make  Eq.~\eqref{eq:indentity} hold. 

Ref.~\cite{schoenberg1938metricsspaces} proved sufficient and necessary conditions to isometrically embed a generic metric space in an euclidean metric space. Applied on the metric Eq.~\eqref{eq:metric}, it would require that the matrix 
\begin{equation}
(\bm{M_i})_{pq} = e^{-\lambda \langle |f(x_i^p, \mathbf{\bar{x}_i}) - f(x_i^q, \mathbf{\bar{x}_i}) | \rangle_\mathbf{\bar{x}_i} } 
\end{equation}
is positive definite. We took the store feature (see Table~\ref{tab:features}) as an example and verified this numerically and found that it is not true. Therefore the store metric space as we defined cannot be isometrically embedded in an Euclidean space.

\begin{figure}
	\centering
	\includegraphics[width=0.5\textwidth]{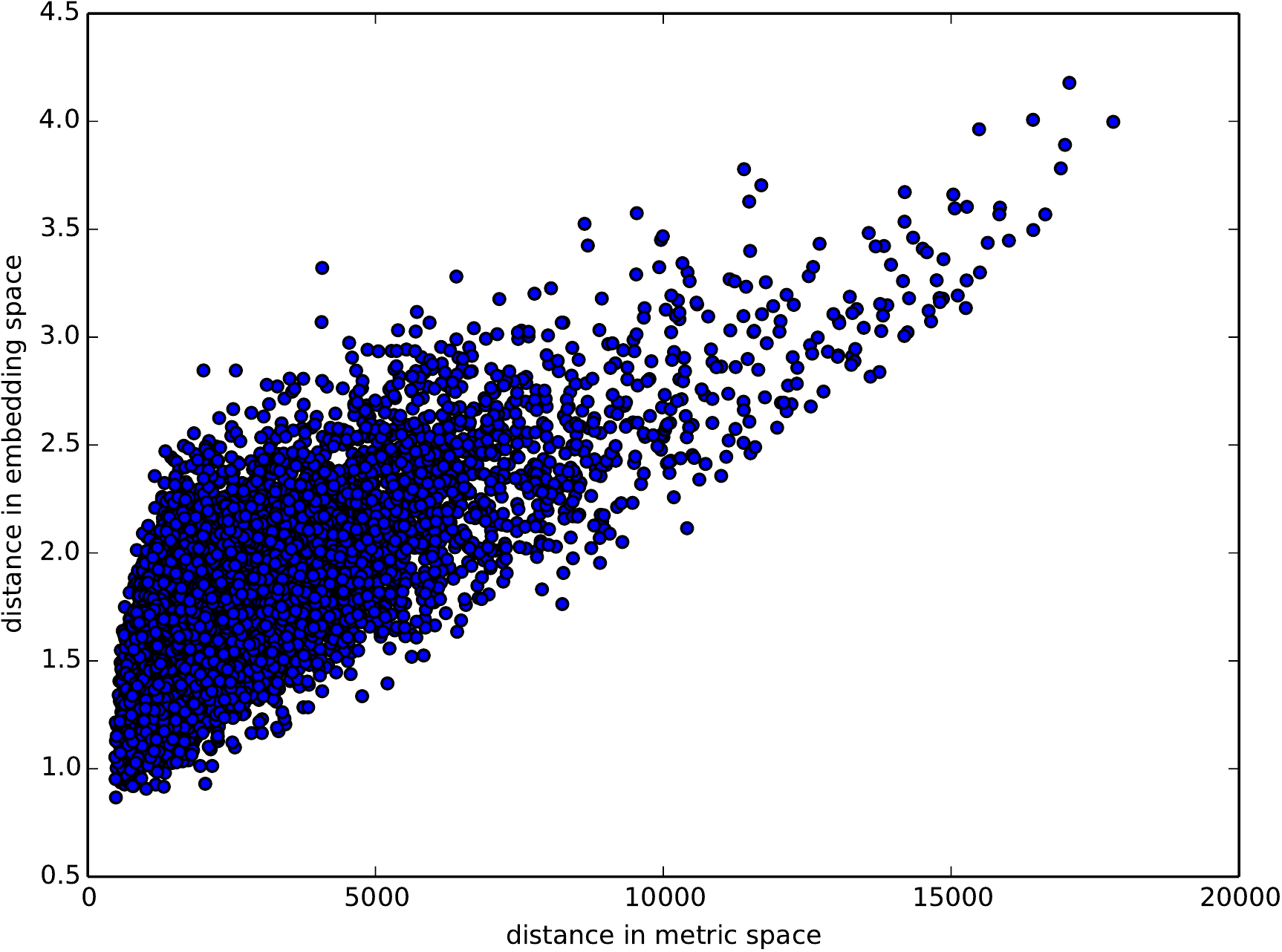}
	\caption{Distance in the store embedding space versus distance in the metric space for 10000 random pair of stores.}
	\label{fig:distance}
\end{figure}

What is the relation of the learned embeddings of a categorical variable to this metric space? To answer this question we plot in Fig.~\ref{fig:distance} the distance between 10000 random store pairs in the learned store embedding space and in the metric space as defined in Eq.~\eqref{eq:metric}. 
It is not an isometric embedding obviously. We can also see from the figure that there is a linear relation with well defined upper and lower boundary. 
Why are there clear boundaries and what does the shape mean? 
Is this related to some theorems regarding the distorted mapping of metric space\cite{abraham2006embedding, Matousek1996distortion}?
How is the distortion related to the embedding dimension $D_i$?
If we apply multidimensional scaling\cite{kruskal1964multidimensional} directly on the metric $d_i$ how is the result different to the learned entity embeddings of the neural network?
Due to time limit we will leave these interesting questions for future investigations.

\section{Experiments}
In this paper we will use the dataset from the Kaggle Rossmann Sale Prediction competition as an example. The goal of the competition is to predict the daily sales of each store of Dirk Rossmann GmbH (abbreviated as 'Rossmann' in the following) as accurate as possible.
The dataset published by the Rossmann hosts\footnote{\url{https://www.kaggle.com/c/rossmann-store-sales/data}} has two parts: the first part is \texttt{train.csv} which comprises about $2.5$ years of daily sales data for $1115$ different Rossmann stores, resulting in a total number of $1017210$ records; the second part is \texttt{store.csv} which describes some further details about each of these 1115 stores.

Besides the data published by the host, external data was also allowed as long as it was shared on the competition forum. Many features had been proposed by participants of this competition.
For example the Kaggle user dune\_dweller smartly figured out the German state each store belongs to by correlating the store open variable with the state holiday and school holiday calendar of the German states (state and school holidays differ in Germany from state to state)\footnote{\url{https://www.kaggle.com/c/rossmann-store-sales/forums/t/17048/putting-stores-on-the-map}}. Other popular external data was weather data, Google Trends data and even sport events dates.

 
In our winning solution we used most of the above data, but in this paper the aim is to compare different machine learning methods and not to obtain the very best result. Therefore, to simplify, we use only a small subset of the features (see Table~\ref{tab:features}) and we do not apply any feature engineering.

\begin{table}
	\begin{tabular}{|c|c|c|c|}
		\hline
		feature & data type & number of values & EE dimension\\
		\hline
		\hline
		store & nominal & 1115  & 10\\
		\hline
		day of week & ordinal & 7 & 6\\
		\hline
		day & ordinal & 31 & 10\\
		\hline
		month & ordinal &12 & 6 \\
		\hline
		year & ordinal & 3 (2013-2015) & 2 \\
		\hline
		promotion & binary & 2 & 1 \\
		\hline
		state & nominal & 12  & 6  \\
		\hline
	\end{tabular}
	\caption{Features we used from the Kaggle Rossmann competition dataset. \textit{promotion} signals whether or not the store was issuing a promotion on the observation date. \textit{state} corresponds to the German state where the store resides. The last column describes the dimension we used for each entity embedding (EE).}
	\label{tab:features} 
\end{table}

The dataset is divided into a $90$\% portion for training, and a $10$\% portion for testing. We consider both a split leaving the temporal structure of the data intact (i.e., using the first $90$\% days for training), as well as a random shuffling of the dataset before the training-test split was applied. 
For shuffled data, the test data shares the same statistical distribution as the training data.
More specifically, as the Rossmann dataset has relatively few features compared to the number of samples, the distribution of the test data in the feature space is well represented by the distribution of the training data.
The shuffled data is useful for us to benchmark model performance with respect to the pure statistical prediction accuracy. 
For the time based splitting (i.e. unshuffled data), the test data is of a future time compared to the training data and the statistical distribution of the test data with respect to time is not exactly sampled by the training data. Therefore, it can measure the model's generalization ability based on what it has learned from the training data.

The code used for this experiment can be found in this github repository\footnote{\url{https://github.com/entron/entity-embedding-rossmann}}.

\subsection{Neural networks}
In this experiment we use both one-hot encoding and entity embedding to represent input features of neural networks. We use two fully connected layers (1000 and 500 neurons respectively) on top of either the embedding layer or directly on top of the one-hot encoding layer. The fully connected layer uses ReLU activation function. The output layer contains one neuron with sigmoid activation function. No dropout is used as we found that it did not improve the result. 
We also experimented with a neural network where the entity embedding layer was replaced with an extra fully connected layer (on top of the one-hot encoding layer) of the same size as the sum of all entity embedding components but the result is worse than without this layer.
We use the deep learning framework Keras\footnote{\url{https://github.com/fchollet/keras}}
to implement the neural network.

As $Sales$ in the data set spans 4 orders of magnitude, we used $\log(Sale)$ and rescaled it to the same range as the neural network output with $\log(Sale) / \log(Sale_{max})$.  Adam optimization method\cite{DBLP:journals/corr/KingmaB14} is used to optimize the networks. Each network is trained for 10 epochs. For prediction we use the average result of 5 neural networks, as an individual neural network showed notable variance.

\subsection{Comparison of different methods}
We compared k-nearest neighbors (KNN), random forests and gradient boosted trees with neural networks. KNN and random forests are tested using the scikit-learn library of python \cite{scikit-learn}, while we use the xgboost implementation of gradient boosted trees \cite{xgboost}. The used model parameters can be found in Table \ref{tab:parameters}. They were empirically found by optimizing the results of the validation set. For the input variables, KNN is fed with one-hot-encoded features, while random forests and gradient boosted trees use the integer coded categorical variables directly. We use $\log(Sales)$ as the target value for all machine learning methods.

As we are using relatively small number of features (7) compared to available training samples (about 1 million) the dataset is not sparse enough for our purpose. Therefore, we sparsified the training data by randomly sampling 200,000 samples out of the training set for benchmarking the models.

Instead of root mean square percentage error (RMSPE) used in the competition we use mean absolute percentage error (MAPE) as the criterion:
\begin{equation}
MAPE = \left\langle  \left\lvert \frac{Sales - Sales_{predict}}{Sales} \right\rvert \right\rangle
\end{equation}
The reason is that we find MAPE is more stable with outliners, which may be caused by factors not included as features in the Rossmann dataset.

\begin{table}
	\begin{tabular}{|c|c|}
	        \hline
	        \multicolumn{2}{|c|}{\textbf{xgboost}} \\
		\hline
		max\_depth & $10$ \\
		\hline
		eta & $0.02$\\
		\hline
		objective & reg:linear\\
		\hline
		colsample\_bytree & $0.7$ \\
		\hline
		subsample & $0.7$ \\
		\hline
		num\_round & $3000$ \\
		\hline
                 \hline
	        \hline
	        \multicolumn{2}{|c|}{\textbf{random forest}} \\
		\hline
		n\_estimators & $200$ \\
		\hline
		max\_depth & $35$\\
		\hline
		min\_samples\_split & $2$\\
		\hline
		min\_samples\_leaf & $1$ \\
		\hline
		\hline
	        \multicolumn{2}{|c|}{\textbf{KNN}} \\
		\hline
		n\_neighbors & $10$ \\
		\hline
		weights & distance\\
		\hline
		p & $1$\\
		\hline
	\end{tabular}
	\caption{Parameters of models used to compare with neural networks. If a parameter is not specified, the default choice of scikit-learn (for random forests and KNN) and xgboost was taken.}
	\label{tab:parameters} 
\end{table}

The results that we obtained can be found in Table~\ref{tab:results_shuffling} and \ref{tab:results_no_shuffling}. 
We can see that neural networks give the best results for non-shuffled data. For shuffled data, gradient boosted trees with entity embedding (see below for an explanation) and neural networks give comparable good results.
Neural networks with one-hot encoding give slightly better results than entity embedding for the shuffled data while entity embedding is clearly better than one-hot encoding for the non-shuffled data. 
The explanation is that entity embedding, by restricting the network in a much smaller parameter space in a meaningful way, reduces the chance that the network converges to local minimums far from the global minimum. 
More intuitively, entity embeddings force the network to learn the intrinsic properties of each of the feature as well as the sales distribution in the feature space.
One-hot encoding, on the other hand, only learns about the sales distribution. A better understanding of the intrinsic properties of the components (features) will give the model an advantage when facing a new combination of the components not seen during training. We expect this effect will be stronger when we add more features, for both shuffled and unshuffled data.

We also used the entity embeddings learned from a neural network as the input for other machine learning methods, that is, we feed the embedded features into other machine learning methods. This significantly improves all the methods tested here as shown in the right columns of the tables.

\begin{table}
	\begin{tabular}{|c|c|c|c|}
		\hline
		method & MAPE & MAPE (with EE) \\
		\hline
		\hline
		KNN & 0.315  & 0.099\\
		\hline
		random forest & 0.167 & 0.089 \\
		\hline
		gradient boosted trees & 0.122 &  0.071 \\
		\hline
		neural network & 0.070 & 0.070 \\
		\hline
	\end{tabular}
 \caption{Comparison of different methods on the Kaggle Rossmann dataset with 10\%  shuffled data used for testing and 200,000 random samples from the remaining 90\% for training.}
	\label{tab:results_shuffling} 
\end{table}

\begin{table}
\begin{tabular}{|c|c|c|c|}
  \hline
  method & MAPE & MAPE (with EE) \\
  \hline
  \hline
   KNN & 0.290 & 0.116\\
   \hline
   random forest & 0.158& 0.108  \\
   \hline
   gradient boosted trees & 0.152 & 0.115   \\ 
   \hline
   neural network & 0.101 & 0.093 \\
   \hline
\end{tabular}
	\caption{Same as Table~\ref{tab:results_no_shuffling} except the data is not shuffled and the test data is the latest 10\% of the data. This result shows the models generalization ability based on what they have learned from the training data.}
  \label{tab:results_no_shuffling}
\end{table}

\subsection{Distribution in the embedding space}
The main goal of entity embedding is to map similar categories close to each other in the embedding space. A natural question is thus how the embedding space and the distribution of the data within it look like. For the following analyses, we used a store embedding matrix of dimension $50$ and trained the network on the full first $90\%$ of data, i.e. we did not apply data sparsification.

\begin{figure}
	\centering
	\includegraphics[width=0.4\textwidth]{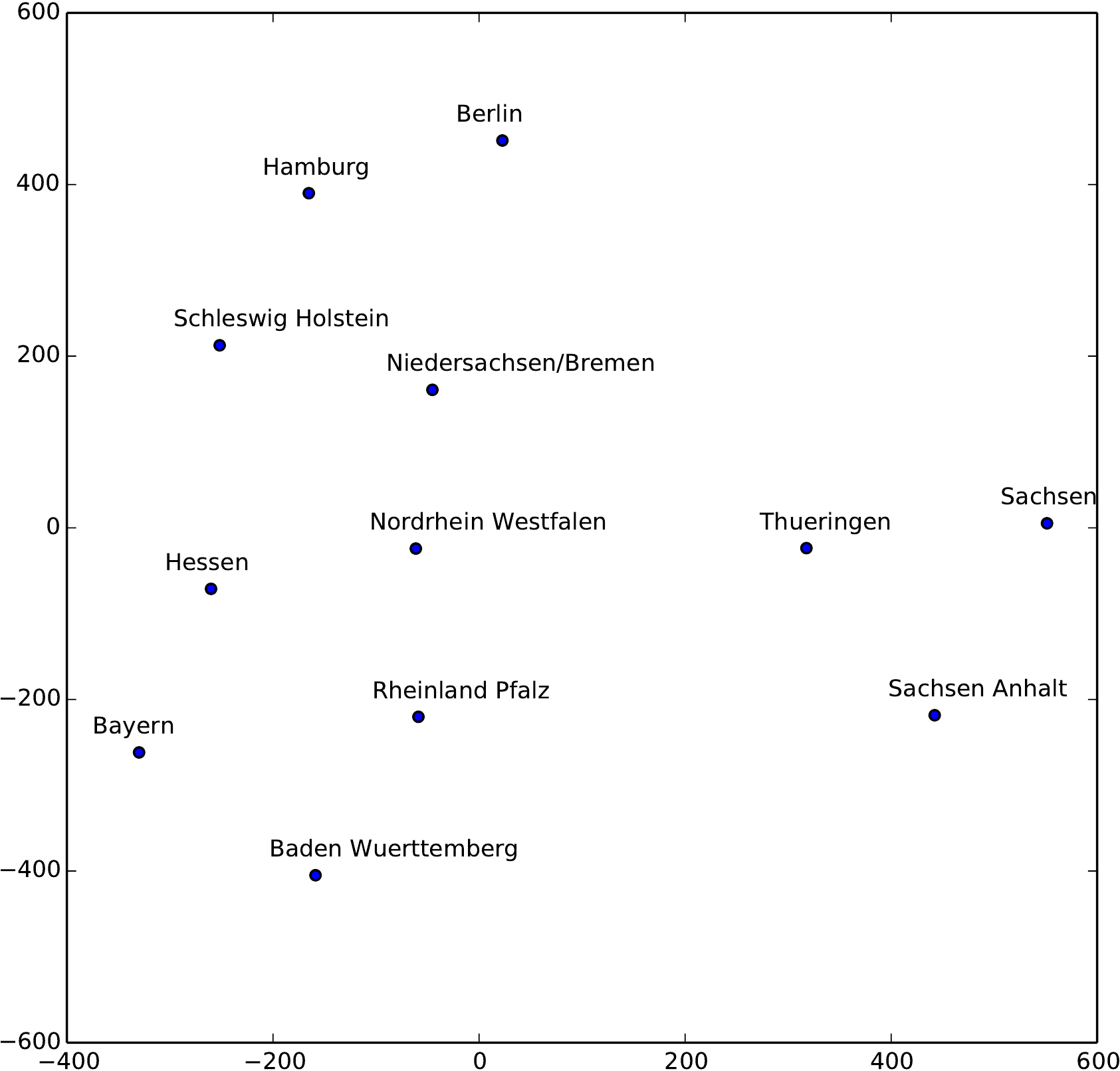}
	\caption{The learned German state embedding is mapped to a 2D space with t-SNE. The relative positions of German states here resemble that on the real German map surprisingly well.}
	\label{fig:state_embedding}
\end{figure}

To visualize the high dimensional embeddings we used t-SNE\cite{van2008visualizing} to map the embeddings to a 2D space.
Fig~\ref{fig:state_embedding} shows the result for the German state embeddings. 
Though the algorithm does not know anything about German geography and society, the relative positions of the learned embedding of German states resemble that on the German map surprisingly well! The reason is that the embedding maps states with similar distribution of features, i.e. similar economical and cultural environments, close to each other, while at the same time two geographically neighboring states are likely sharing similar economy and culture. Especially, the three states on the right cluster, namely \textit{Sachsen}, \textit{Thueringen} and \textit{Sachsen Anhalt} are all from eastern Germany while states in the left cluster are from western Germany. This shows the effectiveness of entity embedding for abductive reasoning. It also shows that entity embedding can be used to cluster categorical data. This is a consequence of entity embedding putting similar values close to each other in an euclidean space equipped with distance measure, on which any known clustering algorithm can be applied. 

\begin{figure}[htb]
    \centering
    \begin{minipage}[t]{0.5\linewidth}
        \centering
        \includegraphics[width=\linewidth]{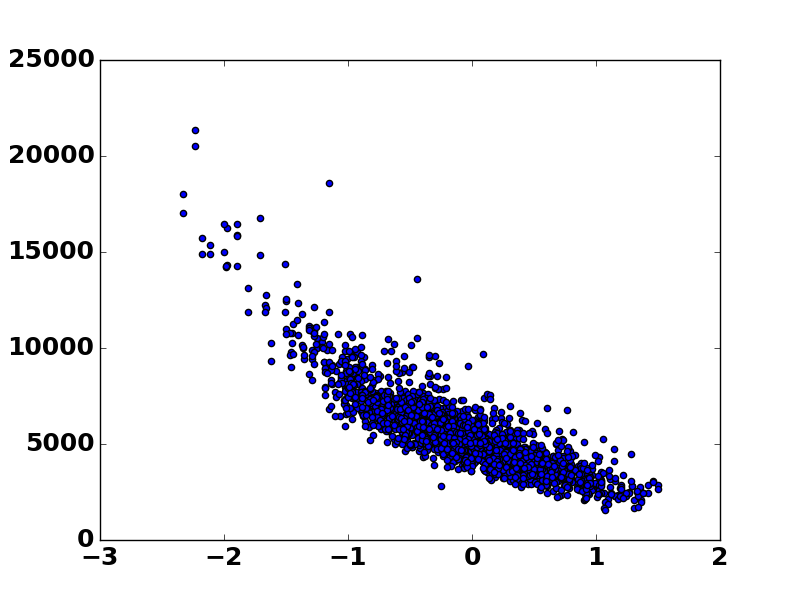}
    \end{minipage}
    \hfill
    \begin{minipage}[t]{0.5\linewidth}
        \centering
        \includegraphics[width=\linewidth]{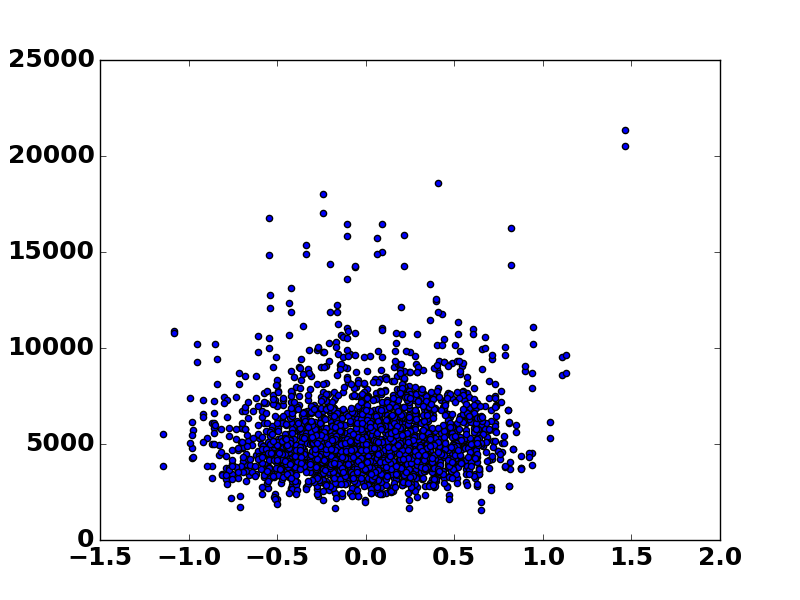}
    \end{minipage}
     \begin{minipage}[t]{0.5\linewidth}
        \centering
        \includegraphics[width=\linewidth]{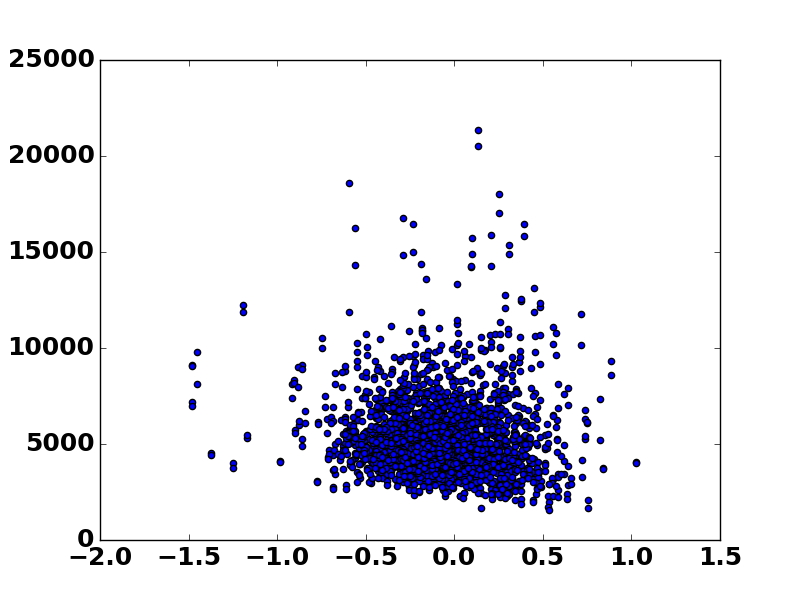}
    \end{minipage}
    \hfill
    \begin{minipage}[t]{0.5\linewidth}
        \centering
        \includegraphics[width=\linewidth]{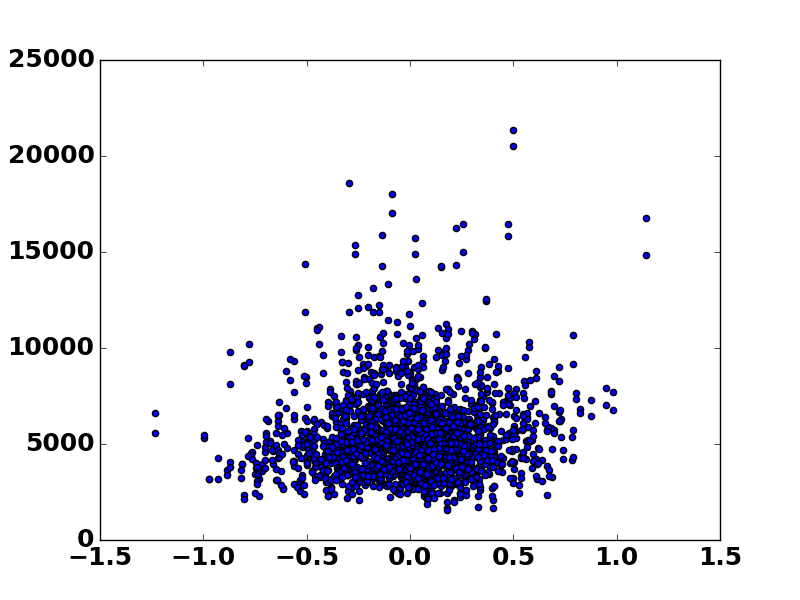}
    \end{minipage}
      \caption{Sales distribution along first principal component (upper left) and second principal component (upper right) of embedded store indices and along two random directions (lower left and right). All $1115$ stores contributed to the plot.}
      \label{fig:sales_storeindex}
\end{figure}

Regarding the sales distribution in entity embeddings, we take entity embedding of the store as an example. Figure \ref{fig:sales_storeindex} shows the sales distribution in the store embedding along its first two principal components and along two random directions. It is apparent from the plot that the sales follows a continuous functional relationship along the first principal component. This allows the neural network to understand the impact of the store index, as stores with similar sales are mapped close to each other. Although the other directions in the subspace have no direct correlation with sales, they are encoding probably other properties of the store and when combined with other features in the deeper layers of the network they could have an impact on the final sales prediction.

\begin{figure}[htb]
    \centering
    \begin{minipage}[t]{0.5\linewidth}
        \centering
        \includegraphics[width=\linewidth]{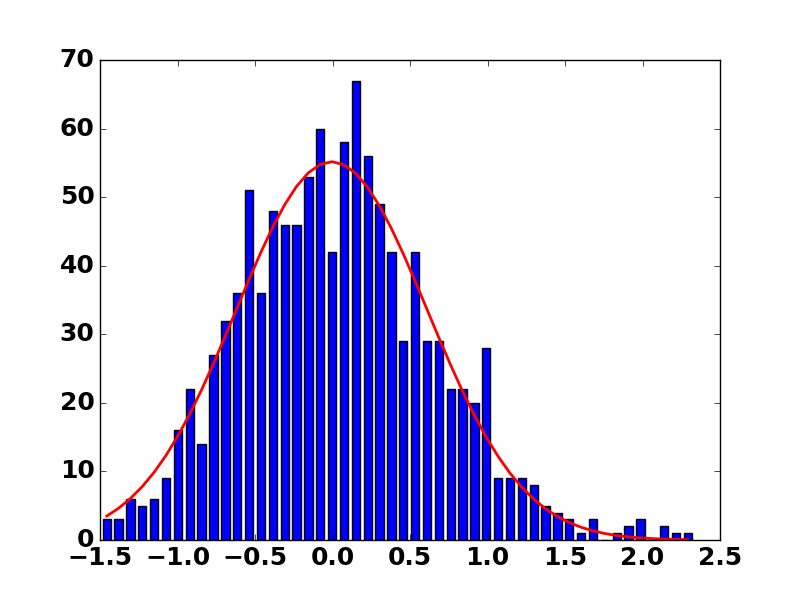}
    \end{minipage}
    \hfill
    \begin{minipage}[t]{0.5\linewidth}
        \centering
        \includegraphics[width=\linewidth]{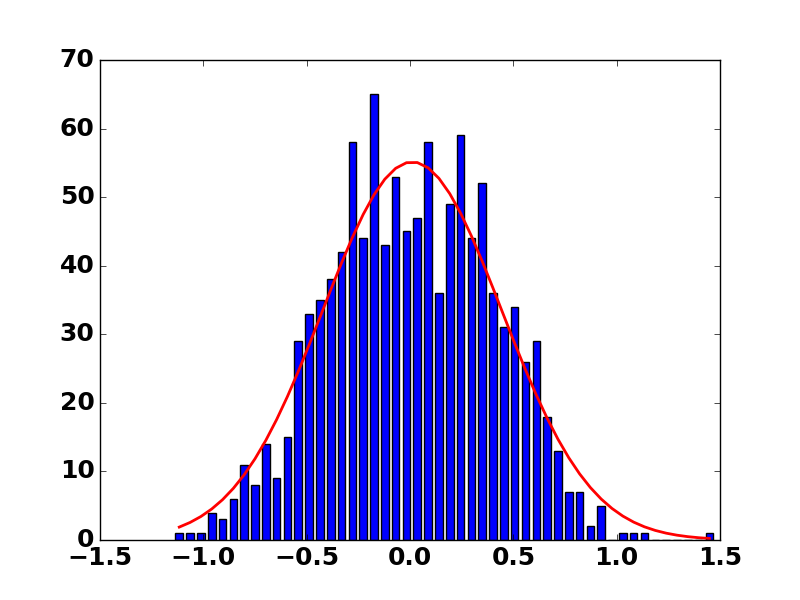}
    \end{minipage}
     \begin{minipage}[t]{0.5\linewidth}
        \centering
        \includegraphics[width=\linewidth]{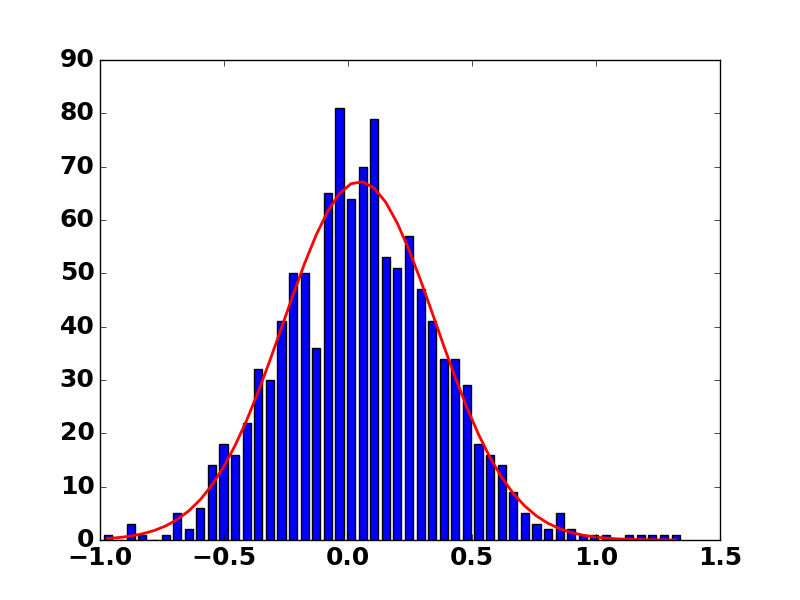}
    \end{minipage}
    \hfill
    \begin{minipage}[t]{0.5\linewidth}
        \centering
        \includegraphics[width=\linewidth]{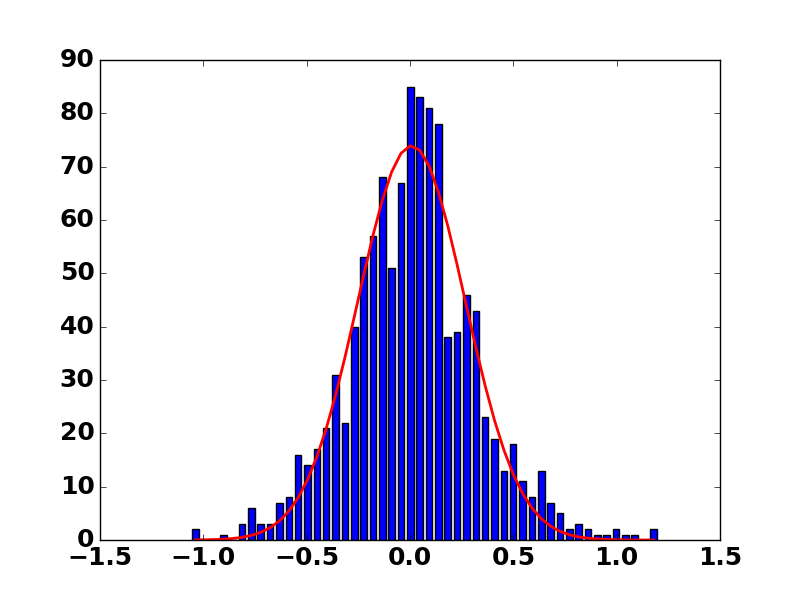}
    \end{minipage}
      \caption{Density distribution of embedded store indices along the first four principal components (from upper left to lower right). The red line corresponds to a gaussian fit. The p-values of the D'Agostino's $K^2$ normality test are all statistically significant, i.e. below $0.05$.}
      \label{fig:density_distribution_storeindex}
\end{figure}

The density distribution of store embedding is visualized in Fig.~\ref{fig:density_distribution_storeindex}, which shows the distribution along the first four principal components. Interestingly, the univariate density along the first principal components is approximately gaussian distributed. However, their joint distribution is not multivariate gaussian, as the Mardia test \cite{mardia1970multigaussian} reveals.   

As can be seen in Fig \ref{fig:ee_and_onehot}, the neural network is fed with the direct product of all the entity embedding subspaces. We also investigated the statistical properties of this concatenated space. We found that there is no strong correlation between the individual subspaces. It is thus sufficient to consider them independently, as we did in this section.

\section{Future Work}
Due to the limitation of time we leave the following points for future explorations:

First of all, entity embedding should be tested with more datasets, in particular datasets with many high cardinality features, where the data is getting sparse and entity embedding is supposed to show its full strength compared with other methods. For some datasets and entity embeddings it could also be interesting to explore the meaning of the directions in the embeddings like those in Eq.~\eqref{eq:word_embedding_examples1} and Eq.~\eqref{eq:word_embedding_examples2}.

Second, we only touched the surface of the relation of entity embedding with the finite metric spaces. A deeper understanding of this relation might also help to find the optimal dimension of the embedding space and how neural networks work in general.

Third, similar methods may be applied to improve the approximation of continuous (i.e. non-categorical), but non-monotone functions. One way to achieve this is by discretizing the continuous variables and transform them into categorical variables as discussed in this paper.

Last, it might be interesting to systematically compare different activation functions of the entity embedding layer. 

\section{Acknowledge}
We thank Dirk Rossmann GmbH to allow us to use their data for the publication. We thank Kaggle Inc. for hosting such an interesting competition. 
We thank Gert Jacobusse for helpful discussions regarding xgboost. 
We thank Neokami Inc. co-founders Ozel Christo and Andrei Ciobotar for their support joining the competition and writing this paper.

\bibliography{references}

\end{document}